\documentclass[letterpaper, 10 pt, conference]{ieeeconf}
\usepackage{todonotes}
\usepackage{subfigure,caption}
\usepackage[utf8]{inputenc} 
\usepackage[T1]{fontenc}    
\usepackage{hyperref}       
\usepackage{url}            
\usepackage{booktabs}       
\usepackage{amsfonts}       
\usepackage{nicefrac}       
\usepackage{microtype}      
\usepackage{xcolor}         

\usepackage{multirow}
\usepackage{chngpage}
\usepackage{diagbox}
\usepackage{cite}
\usepackage{graphicx} 
\usepackage{algorithm}
\usepackage{algpseudocode}
\usepackage{amsmath,amssymb}

\usepackage{xkeyval,xcolor}
\makeatletter
\newlength{\sfp@hseplen}\newlength{\sfp@vseplen}
\define@cmdkey{subfigpos}[sfp@]{pos}[ul]{}
\define@cmdkey{subfigpos}[sfp@]{font}[\small]{}
\define@cmdkey{subfigpos}[sfp@]{vsep}[2\baselineskip]{\setlength{\sfp@vseplen}{\sfp@vsep}}
\define@cmdkey{subfigpos}[sfp@]{hsep}[10pt]{\setlength{\sfp@hseplen}{\sfp@hsep}}
\newcommand{\subfigimg}[3][,]{%
  \setkeys{Gin,subfigpos}{pos,font,vsep,hsep,#1}
  \setbox1=\hbox{\includegraphics{#3}}
  \ifnum\pdfstrcmp{\sfp@pos}{ul}=0
    \leavevmode\rlap{\usebox1}
    \rlap{\hspace*{\sfp@hsep}\raisebox{\dimexpr\ht1-\sfp@vsep}{\sfp@font{#2}}}
    \phantom{\usebox1}
  \else\ifnum\pdfstrcmp{\sfp@pos}{ur}=0
    \leavevmode\usebox1
    \llap{\raisebox{\dimexpr\ht1-\sfp@vsep}{\sfp@font{#2}}\hspace*{\sfp@hsep}}
  \else\ifnum\pdfstrcmp{\sfp@pos}{lr}=0
    \leavevmode\usebox1
    \llap{\raisebox{\sfp@vsep}{\sfp@font{#2}}\hspace*{\sfp@hsep}}
  \else\ifnum\pdfstrcmp{\sfp@pos}{lr2}=0
    \leavevmode\usebox1
    \llap{\raisebox{8pt}{\sfp@font{#2}}\hspace*{\sfp@hsep}}
  \else
    \leavevmode\rlap{\usebox1}
    \rlap{\hspace*{\sfp@hseplen}\raisebox{\sfp@vsep}{\sfp@font{#2}}}
    \phantom{\usebox1}
  \fi\fi\fi
}

\newtheorem{definition}{Definition}

\begin{document}

\title{\LARGE \bf Sequential Neural Barriers for Scalable Dynamic Obstacle Avoidance}

\author{Hongzhan Yu, Chiaki Hirayama, Chenning Yu, Sylvia Herbert, Sicun Gao}

\maketitle
\thispagestyle{empty}
\pagestyle{empty}

\begin{abstract}
There are two major challenges for scaling up robot navigation around dynamic obstacles: the complex interaction dynamics of the obstacles can be hard to model analytically, and the complexity of planning and control grows exponentially in the number of obstacles. 
Data-driven and learning-based methods are thus particularly valuable in this context. 
However, data-driven methods are sensitive to distribution drift, making it hard to train and generalize learned models across different obstacle densities. We propose a novel method for compositional learning of Sequential Neural Control Barrier models (SN-CBFs) to achieve scalability. Our approach exploits an important observation: the spatial interaction patterns of multiple dynamic obstacles can be decomposed and predicted through temporal sequences of states for each obstacle. Through decomposition, we can generalize control policies trained only with a small number of obstacles, to environments where the obstacle density can be 100x higher. We demonstrate the benefits of the proposed methods in improving dynamic collision avoidance in comparison with existing methods including potential fields, end-to-end reinforcement learning, and model-predictive control. We also perform hardware experiments and show the practical effectiveness of the approach in the supplementary video. 
\end{abstract}


\section{Introduction}

Dynamic obstacle avoidance poses longstanding challenges for mobile robots. Consider the case of autonomous driving in populated areas:
the ego-robot needs to quickly predict the movement of the pedestrians and infer control actions that can avoid collision accordingly, while maintaining progress towards its goal. 
Existing approaches typically use known dynamics of both the obstacles (i.e. pedestrians) and the ego-robot to compute control actions, using methods such as artificial potential fields
(APFs)~\cite{potential86}, dynamic windows~\cite{fox1997dynamic}, and model-predictive control (MPC)~\cite{kouvaritakis2016model,ji2016path}. 
Control barrier functions (CBFs)~\cite{Ames2019CBFOverview,Singletary2020APFCBF} provide a new approach~\cite{xiao2019control, huang2020switched, breeden2021robust} that combines the benefits of potential fields and MPC. CBFs reduce the complexity of online optimization by enforcing a value landscape that maintains forward invariance of safe behaviors of the ego-robot. They still require full knowledge of the dynamics of the system, and can be hard to design in complex environments. CBFs can also encounter the issue of ``freezing robots" when used for ensuring collision avoidance with multiple dynamic obstacles~\cite{Fridovich2020HumanJournal,freezingrobot}. 


\begin{figure}[th!]
    \centering
    \includegraphics[width=0.28\textwidth]{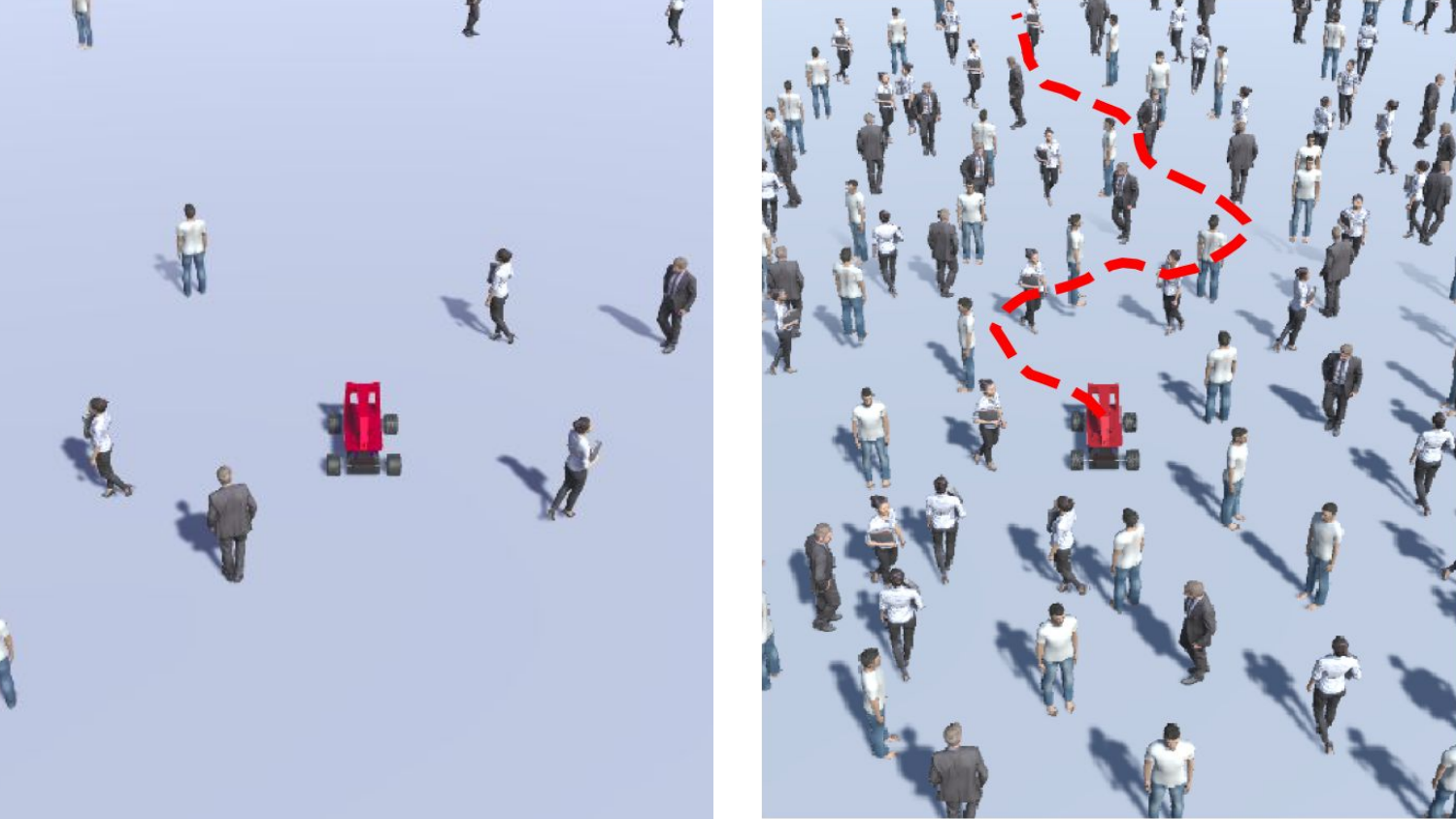}
    \includegraphics[width=0.17\textwidth]{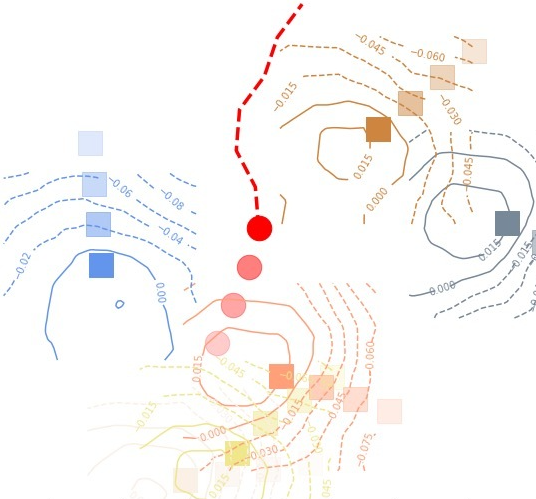}
    \vspace{2mm}
    
    \includegraphics[width=0.47\textwidth]{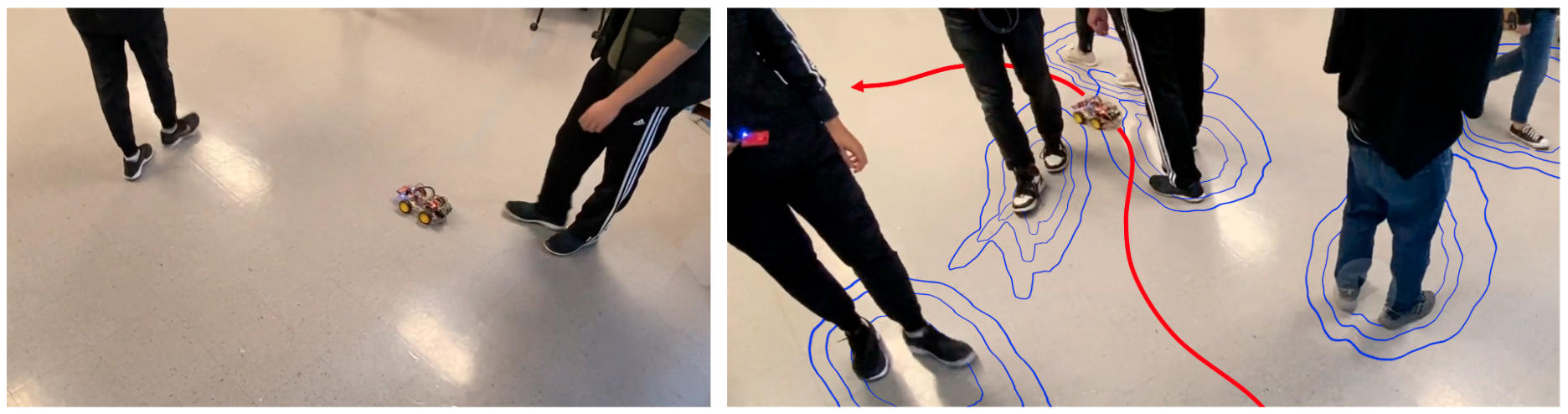}
    \caption{\small (Top Row) Illustration of our goal of generalizing from training environment with a small number of dynamic obstacles to test environments that are much more densely populated. The top right figure shows a high-level sketch of our methods. The red dot in the middle is the ego-robot, and the adjacent squares are the dynamic obstacles. The contours indicate the level sets of the learned SN-CBF models for each dynamic obstacle. The ego-robot iteratively computes control actions based on the SN-CBF values to achieve collision avoidance. (Second Row) Illustration of the hardware experiment setting. The car in the middle is the ego-robot, which is trained with only 2 pedestrians, and then directly deployed around 6 pedestrians and achieves collision avoidance. Level curves of the SN-CBF models and the path of the car are also illustrated. }
    \vspace{-5mm}
    \label{fig::unity}
    \end{figure}

A major difficulty with dynamic obstacles, such as humans, is that the analytic modeling of their dynamics is inherently hard~\cite{rudenko2020human,Kruse2013HumanAwareNav}. For specific applications, it is often viable to collect data to train black-box models that make accurate predictions, in the form of neural networks~\cite{willard2020integrating} or Gaussian processes~\cite{kocijan2016modelling}. 
However, they have two drawbacks:

\noindent{1)} 
{\em Hard to Scale and Generalize.} The interaction patterns of the dynamic obstacles grow exponentially in the number of obstacles, which affects both training and inference. 
Training is expensive because of the need to sample the combinatorial space of possible patterns of all dynamic obstacles, and 
distribution drift becomes a major challenge~\cite{nair2018overcoming,yang2020reinforcement}. 
If we train a control policy in an environment with a small number of pedestrians, then the policy will struggle in environments with a large number of pedestrians that exhibit a very different distribution in the obstacle dynamics (Figure~\ref{fig::unity}).

\noindent{2)} 
{\em Hard to Optimize for Predictive Control.} Although high-capacity learning-based models can fit the collected data with high accuracy, they are extremely nonlinear functions that can not be easily used to form online optimization problems, such as for MPC. They can be used through forward-unrolling and sampling, which often becomes inefficient and unreliable for real-time inference of the control actions. 


In this paper, we propose a new approach to alleviate both limitations of learning-based methods for dynamic obstacle avoidance at scale. 
The key technique is based on the following observation: the \textit{collective} dynamics of the dynamic obstacles can be approximately inferred from the sequential patterns in the trajectories of each \textit{individual} obstacle. 
For instance, when we observe that one pedestrian is slowing down or changing directions, it is most likely because of other pedestrians or obstacles nearby. That allows us to directly infer the next state of the pedestrian, without the need of 
explicitly using the spatial information of the other obstacles. 
In this way, the collective spatial interaction dynamics of a group of dynamic obstacles can be inferred by aggregating the predictions from the sequential patterns of each obstacle. Such inference can be hard to formulate analytically, but high-capacity neural network models may capture such implicit patterns through data. We will first examine the validity of such decomposition in detail in Section~\ref{section::validity_exp}, and show that it is central to achieving scalable modeling and control. 

\begin{figure}[t!h]
\centering
\subfigimg[width=0.45\textwidth, pos = lr, font=\LARGE]{(a)}{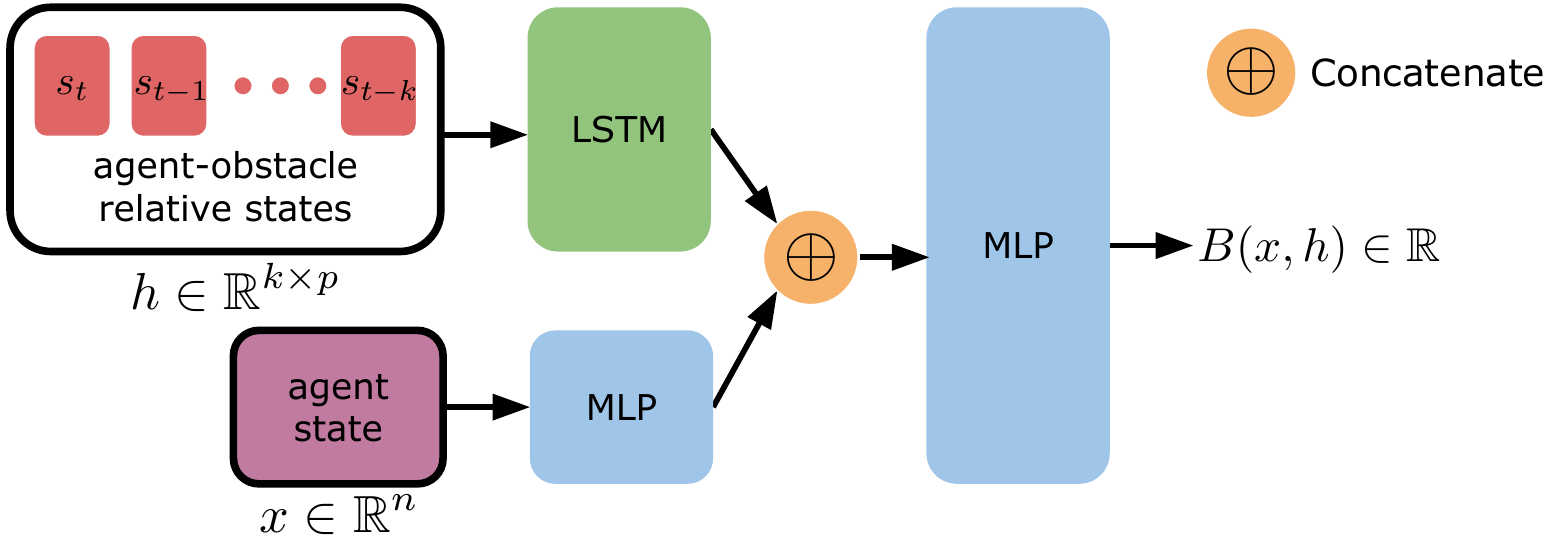}
\hspace{.2cm}
\subfigimg[width=0.25\textwidth, pos = lr2, font=\LARGE]{(b)}{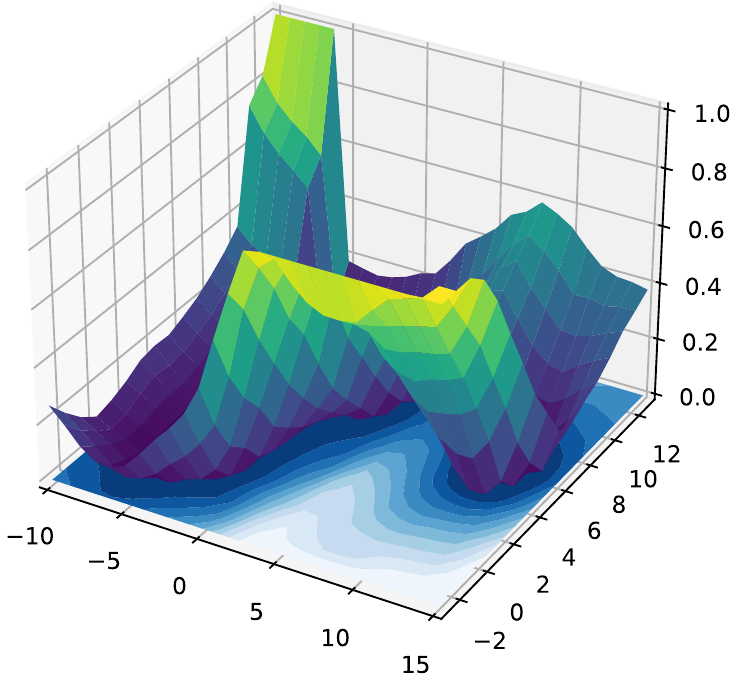}
\subfigimg[width=0.22\textwidth, pos = lr2, font=\LARGE]{(c)}{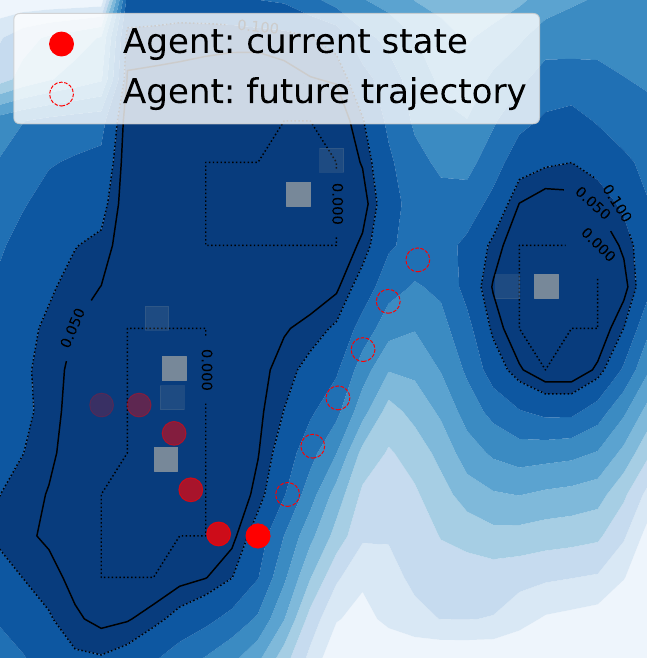}
\caption{\small (a): Neural network architecture for the SN-CBF models. (b): 3D plot of the combined value landscape, where the zero-level sets form the safety barrier. (c): 2D plot of the aggregated value landscape for multiple obstacles (colored in grey). 
The agent trajectory (colored in red) demonstrates the safe controls inferred after aggregating the SN-CBF values from each obstacle.}
\vspace{-5mm}
\label{fig::architecture}
\end{figure}

Given the benefits of compositional learning with sequential models, we propose the design of sequential neural control barrier functions (SN-CBFs) to achieve compositional learning and inference for scalable dynamic collision avoidance. Note that the design does not rely on the direct use of sequential models to predict the movement of the obstacles.
Instead, by learning SN-CBF models, we can directly infer safe control actions for the current state of the ego-robot, without the need of unrolling the complex predictive models.
Moreover, the highly nonlinear SN-CBF models can produce value landscapes
that are significantly more complex than manually-designed simpler forms of potential fields or barrier functions, as illustrated in Figure~\ref{fig::architecture}(c).  In this way, the SN-CBF alleviates well-known issues, such as the narrow-corridor effects in APF, and can be used on ego-robots with highly nonlinear dynamics (details in Section~\ref{section::experiment}). 

Importantly, although the SN-CBF models are first applied to each dynamic obstacle individually, the control action is always computed after aggregating the value landscapes for {\em all obstacles} at every step. As illustrated in Figure~\ref{fig::architecture}(b-c)
, we aggregate the SN-CBF values from all obstacles into one unified landscape to infer the control actions for the ego-robot (red dot in the figure). Doing so alleviates the common issue of ``freezing robots,'' where simply computing the ego-robot control with respect to each dynamic obstacle can easily lead to conflicting control decisions~\cite{freezingrobot}. In contrast, every control action that we successfully obtain from SN-CBF models avoids all obstacles simultaneously. 
We analyze the performance of our method in Section~\ref{section::experiment}, showing that it maintains a significantly lower failure rate compared to existing methods, especially as the obstacle density increases. 


We will describe our contributions in the following order. We will first formalize and evaluate the sequential decomposability of the collective interaction patterns of dynamic obstacles in Section~\ref{section::validity_exp}. We will then describe the model-free learning procedures for the SN-CBF models in Section~\ref{section::learning}, and then the online inference procedures in Section~\ref{section::planning}. We evaluate the proposed methods in simulation environments and hardware experiments in 
Section~\ref{section::experiment}. We demonstrate scalable performance in  collision avoidance that generalizes well from sparse to dense environments. We analyze how the new methods can address common issues in potential fields, reinforcement learning, and model-predictive approaches.

\section{Related Work}


\noindent{\bf Dynamic Collision Avoidance.} Existing methods for dynamic collision avoidance typically require the known dynamics of both the ego-robot and the obstacles.
Artificial potential fields (APFs) methods
~\cite{potential86, Barraquand1992APF, Ge2002APF} design repulsive/attractive potential fields and use the gradient of this function to inform a feedback controller. 
They typically require that the ego-robot and the obstacles have known and simple dynamics such that the gradient directions can be directly followed. 
Under such assumptions, APFs can be used at large scales~\cite{Henderson1971Fluid, Rudenko2020HumanPredictionSurvey},
 but capturing human movements using simple potential fields requires strong assumptions. 
Model-predictive control (MPC)~\cite{kouvaritakis2016model, ji2016path} is another main framework for dynamic collision avoidance. It formulates online optimization problems that involve unrolling the system dynamics of both the ego-robot and the obstacles over bounded time horizons, to compute optimizing control actions.
MPC can have high computational complexity, and additional efforts are required for handling disturbances and modeling error~\cite{bemporad1999robust}. 
The dynamic window approach~\cite{fox1997dynamic} is a special form of MPC that reduces the search space to admissible controls of the ego-robot, which has also been extended to use learned dynamics models based on the collected data~\cite{stenman1999model,pan2011model}. 
The prediction error can quickly accumulate, and we will show the advantage of our proposed methods compared with such methods in the experiments.

\noindent{\bf Learning-based Approaches.}
Deep reinforcement learning (DRL) approaches have been proposed for dynamic obstacle avoidance in many forms, 
including CADRL~\cite{chen2017socially}, MRCA~\cite{long2018towards}, and GA3C-CADRL~\cite{everett2021collision}. 
These methods focus on formulating the avoidance problems as Markov Decision Processes (MDPs) or Partially-Observed MDPs (POMDPs) to perform model-free learning of the control policies. 
CADRL~\cite{chen2017socially} encodes social interactions into reward shaping for RL training to achieve safe navigation in pedestrian-rich environments. 
MRCA~\cite{long2018towards} performed collision avoidance through information on LIDAR measurements without directly detecting the dynamics objects.
GA3C-CADRL~\cite{everett2021collision} introduced sequential models to support a varying number of pedestrian states. 
GCBF-MBPO~\cite{ma2021model} proposed model-based enhancement to achieve faster training. 
In general, existing DRL methods are sensitive to distribution drift and lack generalizability from sparse to dense environments. 
We will compare with DRL baselines in the experiment section. 



\noindent{\bf Control Barrier Functions.} 
Control barrier functions (CBFs)~\cite{Ames2019CBFOverview,Singletary2020APFCBF} impose (typically manually-designed) value landscapes to ensure forward invariance of the safe set with control actions computed by efficient online optimization (as quadratic programs). 
While well-designed CBFs can provide formal guarantee for control systems with static obstacles and known dynamics, its direct application in dynamic obstacle avoidance~\cite{Gurriet2018CBFActiveSetInvariance,Nguyen2016CBFSteppingStones1} has several challenges. 
Applying CBFs between every pair of agents lead to feasibility issues where avoiding one agent inevitably leads to collisions with another,
while synthesizing valid CBFs for arbitrary numbers of agents is challenging. To mitigate the issue of feasibility and scalability, several recent works have proposed compositional CBFs. They can be constructed through temporal logic~\cite{Huang2020MultiAgentCBF,Srinivasan2018MulitAgentCBF}, or piecewise CBFs~\cite{DBLP:conf/cdc/WangAE16,DBLP:journals/trob/WangAE17,DBLP:journals/csysl/GlotfelterCE17,DBLP:conf/ccta/GlotfelterCE18}.
Learning-based approaches have been introduced for constructing CBFs from sensory data with linear functions~\cite{saveriano2019learning}, support vector machines~\cite{Srinivasan2020LearnCBFSupervisedML}, and neural networks~\cite{Jin2020,Taylor2020LearningCBF, Robey2020LearnCBFExpertDemonstrations}. The work in \cite{macbf, Dawson2021CBFNN} shows the benefits of jointly learning CBFs as safety certificates and the control policies. 
The work in 
\cite{macbf} uses neural network CBFs to achieve safe decentralized control in multi-agent systems, assuming known nonlinear dynamics. 
The work in~\cite{ma2022learning} generalizes CBFs to new configurations of static obstacles, while we consider generalization from sparse to dense environments of dynamic obstacles. We focus on learning sequentially decomposable value landscapes, instead of reactive control policies, for dynamic obstacles without known dynamics, such that safe control action can be efficiently performed online at scale.

\section{Preliminaries}

We consider ego-robots with underlying dynamics $\dot{x}(t)=f(x(t), u(t))$ where $x(t)$ takes values in an $n$-dimensional state space $X\subseteq \mathbb R^n$, $u(t)\in U\subseteq \mathbb{R}^m$ is the control vector, and $f:X\times U \rightarrow \mathbb{R}^{n}$ is a Lipschitz-continuous vector field.
We allow $f$ to be generally nonlinear and {\em not control-affine}, unlike typically assumed in CBF methods. 
Safety properties, such as collision avoidance, can be specified by declaring an unsafe region of the state space $X_u\subseteq X$. We say the system is safe if none of its trajectories intersects with $X_u$. 

To ensure safety properties of a system, we can construct a {\em forward invariant set} for the system that is disjoint from the unsafe set. We say a subset of the state space $\mathrm{Inv}\subseteq X$ is forward invariant for the agent under control, if for any initial state $x(0)\in \mathrm{Inv}$ and any $t\geq 0$, we have $x(t)\in \mathrm{Inv}$. Namely, any trajectory that starts in the invariant $\mathrm{Inv}$ stays in $\mathrm{Inv}$ forever. Consequently, a system is safe if we can find a forward invariant set $\mathrm{Inv}$ such that $\mathrm{Inv}\cap X_u=\emptyset$. 
CBFs are scalar functions whose zero-superlevel set is a forward invariant set in the safe region of the space, and whose spatial gradients can be used to enforce this invariance.
\begin{definition}[Control Barrier Functions~\cite{Ames2019CBFOverview}] \label{def:barrier_func}
Consider a dynamical system defined by vector field $f:X\times U \rightarrow X$ where $X\subseteq \mathbb{R}^n$ is the state space and $U\subseteq \mathbb{R}^m$ the control space. 
Let $B: X\rightarrow \mathbb{R}$ be a continuously differentiable function with zero-superlevel set $\mathcal{C}=\{x\in X: B(x)\geq 0\}$. 
We say $B$ is a control barrier function, and $\mathcal{C}$ is forward invariant, if for any state $x \in X$:
\begin{equation} \label{equ:barrier_function}
\begin{aligned}
\max_{u\in U} \dot B(x) = \max_{u\in U}\langle \nabla B(x), f(x, u)\rangle \geq -\alpha (B(x))
\end{aligned}
\end{equation}
Here $\dot B(x)$ is the Lie derivative of $B$. $\langle \cdot,\cdot \rangle$ denotes inner product. ${\alpha} (\cdot)$ is an extended class-$\mathcal{K}_{\infty}$ function. 
We often choose $\alpha(B(x))=\kappa B(x)$ for some parameter $\kappa \in\mathbb{R}^+$.

\end{definition}

In this paper, we consider model-free training in stochastic environments, and do not attempt to globally satisfy the standard CBF conditions~(\ref{equ:barrier_function}). 
Instead, we encode the conditions as loss functions, and use the idea of CBFs to reduce collision rate with statistical evaluation of its effectiveness, rather than to prove the complete absence of collision. 




\section{Compositional Sequential Modeling of Spatial Interaction Dynamics}
\label{section::validity_exp}

\begin{figure*}[t!]
\centering
\includegraphics[width=0.18\textwidth]{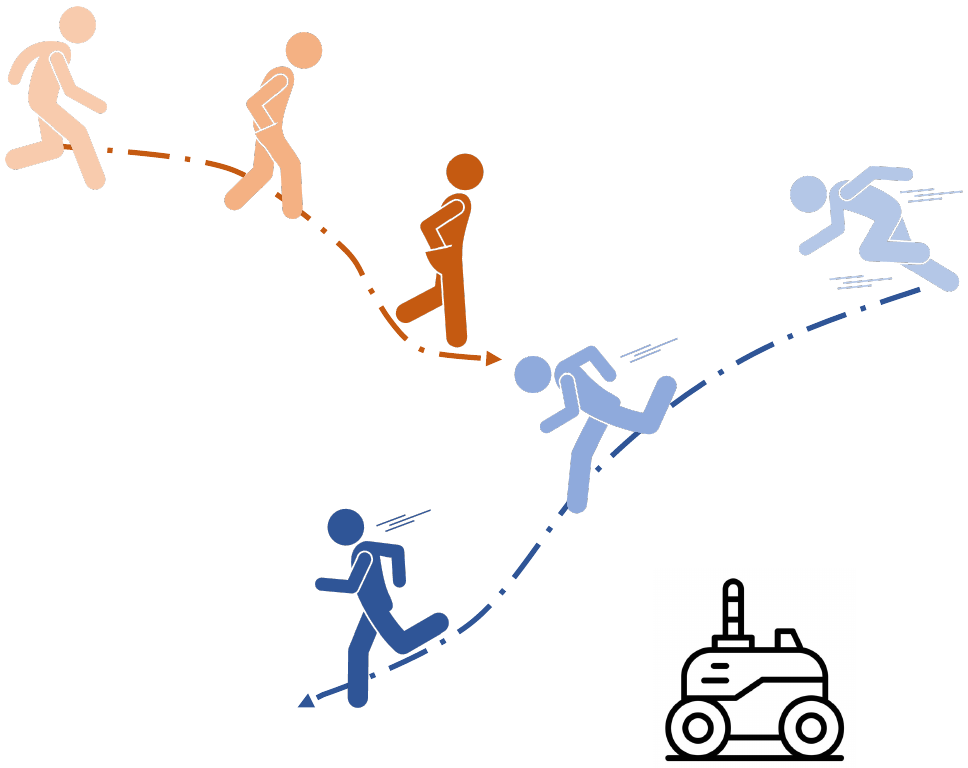}
\includegraphics[width=0.58\textwidth]{plots/validity_exp/validity_exp_example_N15.pdf}
\includegraphics[width=0.22\textwidth]
{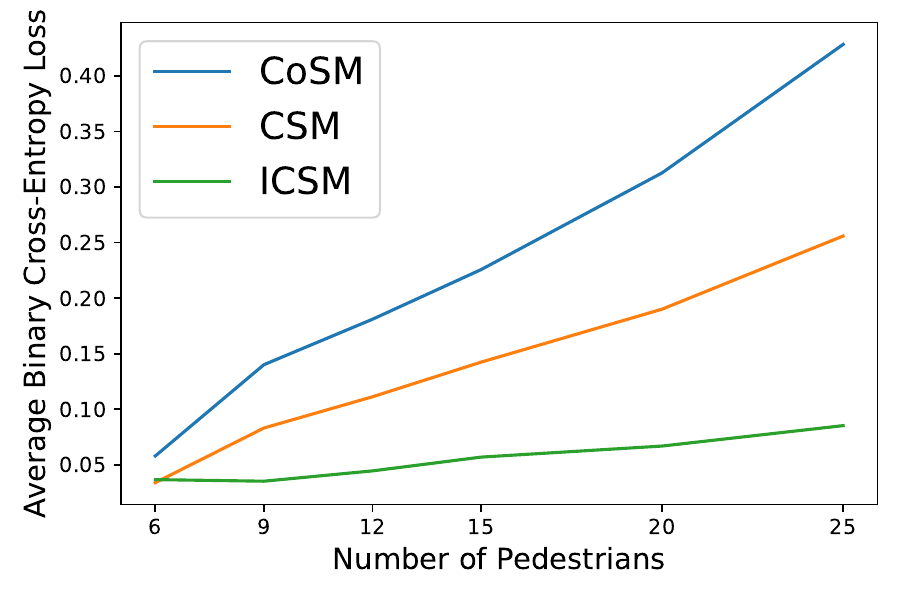}
\caption{\small (Left) Sequential decomposition of pedestrian dynamics. (Middle) Comparing the test performance of different approaches, applying models trained in spare environments to a denser environment. The proposed approach ICSM gives much better prediction, measured by the similarity with the ground truth. (Right) The prediction error grows as density increases with ICSM maintaining the lowest error and outperforming the others. }
\vspace{-6mm}
\label{fig::validity_exp_overview}
\end{figure*}

Our approach builds on a key observation: the collective dynamics of the dynamic obstacles, such as how a group of pedestrians interacts with each other, can be approximately inferred by aggregating the prediction for each individual obstacle based on the sequential patterns in their own trajectories. We now formalize and evaluate this claim first. 

Suppose the state of each dynamic obstacle can be fully described as a vector in $\mathbb{R}^q$, 
such that for $m$ such obstacles their joint state is $h=(h_1, ..., h_m)\in\mathbb{R}^{mq}$, where $h_i$ is the state vector for the $i$-th obstacle. The spatial interaction dynamics of such $k$ obstacles is the vector field defined over the space of the joint states. To differentiate the dynamics of the obstacles from the dynamics of the ego-robot, we write it as a discretized mapping over consecutive states as \[G:\mathbb{R}^{mq}\rightarrow\mathbb{R}^{mq}, h^{(t+\Delta t )}=G(h^{(t)})\] 
where $h^{(t)}$ is the joint state of all obstacles at time $t$ and $\Delta t$ is a small time step. The difficulty with modeling $G$ through sampling state pairs in the joint state space $\mathbb{R}^{mq}$ is two-fold. First, the sample complexity over the $\mathbb{R}^{mq}$ grows exponentially in the number of obstacles $m$. Second, the dynamics and distribution learned for any fixed $m$ may not be applicable to a different $m'$ number of obstacles. 

We assert that for each individual $h_i$, their dynamics should have certain regularity in the sense that they typically react to similar observations of other agents in the same way, which is identified by the state trajectories of $h_i$ itself. For instance, in Figure~\ref{fig::validity_exp_overview}(Left), by observing that the agent on the left (colored in orange) slows down quickly, we can infer that its immediate next state should continue to slow down. We know this without directly observing the agent's state on the right (colored in blue). The same principle applies to this other agent: from its sequence of states, we can infer that it is picking up speed while curving a little bit to avoid another agent. Thus, by only observing the two separate sequences of each agent, we can aggregate the individual predictions, and infer their joint next state, which bypasses the need to learn the collective state transition. Formally:
\begin{definition}[Sequential Decomposability]
Let $\{h_i\}_{i\in [m]}$ be the state vectors describing $m$ dynamic obstacles,
where each $h_i\in\mathbb{R}^q$. Let the collective dynamics of the joint state be defined by $G:\mathbb{R}^{qm}\rightarrow\mathbb{R}^{qm}$. We say $G$ is sequentially decomposable in $k\in\mathbb{Z}^+$ steps up to $\varepsilon\in\mathbb{R}^+$, if there exists $\hat G_i^k:(\mathbb{R}^{q})^k\rightarrow \mathbb{R}^{q}$ for each $h_i$ of the form
\[\hat h_i^{(t+\Delta t)}=\hat G_i^k(h_i^{t},h_i^{(t-\Delta t)},...,h_i^{(t-k\Delta t)})\]
where $t\geq k\Delta t$, such that 
\[\|(\hat h_1^{(t+\Delta t)},...,\hat h_m^{t+\Delta t})-G(h_1^{(t)},...,h_m^{(t)})\|_{\infty}\leq \varepsilon.\]
In words, the approximate prediction of the next state for all obstacles predicted by $\hat G_i^k$ is within $\varepsilon$-error from the ground truth interaction dynamics $G$ in the max norm. Importantly, $\hat G_i^k$ only considers the states of an individual $h_i$ as its inputs. 
\end{definition}

While we can not directly prove the sequential decomposability without precise analytic models of the dynamic obstacles, we can empirically evaluate its validity for given systems. For pedestrian dynamics, we simulate the interaction dynamics of the pedestrians using the widely-adopted ORCA model~\cite{ORCA}. We train the sequential models of individual obstacles to perform coordinate-wise safety classification in an environment with 6 pedestrians and test it in higher numbers of obstacles, and compare with baselines as follows. 
We experiment with several designs of generalizing neural network models from sparse to dense environments. 
First, we consider the approach of using a permutation-invariant encoder over pedestrians with sequences of all states, so that it can be applied to arbitrary number of pedestrians, but can not handle the inherent distribution drift when the obstacle density changes from training to tests. We call this first design the \textit{Collective Sequential Model (CoSM)}. 
The second design, called \textit{Compositional Sequential Model (CSM)} uses a sequential model with individual pedestrian states but does not condition the learning with interaction among pedestrians. This design achieves better prediction and generalization. The third design, named \textit{Interaction-based Compositional Sequential Model (ICSM)}, corresponds to our main approach in SN-CBF, taking into account both the sequential data and the interaction of the nearby agents. 
Figure~\ref{fig::validity_exp_overview} (Middle-Right)
demonstrates that the sequential decomposition plus interaction of nearby obstacles produces the best accuracy and generalizability. 
Note that the SN-CBF model will not directly predict the next states of the obstacles, but will generate value landscapes that aim to capture both state sequence patterns from individual obstacles, and also the implicit interaction patterns of the nearby obstacles exhibited in training data.

\section{Training Procedures for SN-CBF Models}\label{section::learning}






\subsection{Model Architecture}


We design the SN-CBF models to allow an implicit parameter space $H\subseteq \mathbb{R}^{k\times q}$, where $H$ contains length-$k$ sequences of the obstacle states, relative to the ego-robot, where each relative state $h^{(t)}\in \mathbb{R}^{q}$. 
The SN-CBF model can then be conditioned on such sequential information, and still produce scalar values over the ego-robot state $x\in X\subseteq \mathbb{R}^n$. Namely, the models are functions $B: X\times H\rightarrow \mathbb{R}$, with $B(x,h)$ giving a scalar value on the robot state $x$ given the observation $h^{(t)},...,h^{(t-k\Delta t)}$ of the obstacle's state sequence. 
SN-CBF models are constructed using the architecture shown in Figure~\ref{fig::architecture}(a), where we encode $h$ with a standard long short-term memory (LSTM) neural network for handling sequential inputs~\cite{hochreiter1997long}, and the ego-state $x$ is embedded through a multilayer perceptron (MLP). We concatenate the encoded vectors as $d$, and feed $d$ to another MLP that computes the CBF value $B(x, h) \in \mathbb{R}$. This architecture is important for the generalizability of the learned model. 
\subsection{Training Procedures}

\begin{figure}[t!]
\centering
\includegraphics[width=0.23\textwidth]{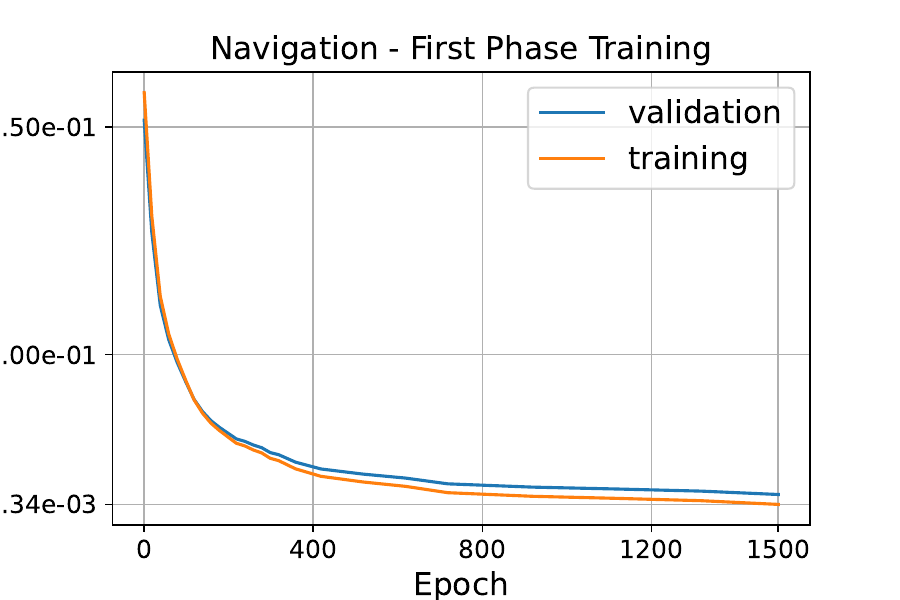}
\includegraphics[width=0.23\textwidth]{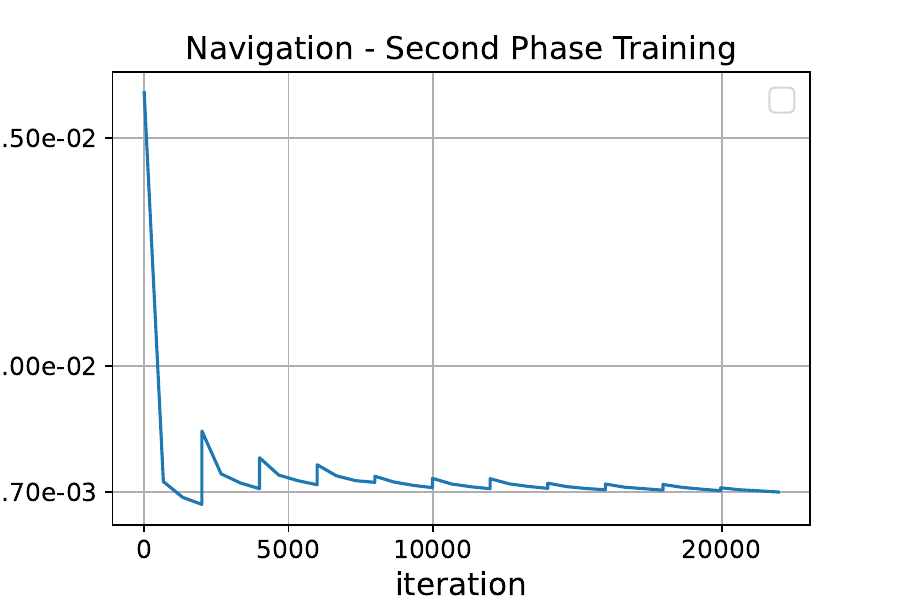}
\caption{\small The learning curves for two phases in the navigation environment. Note that the initial loss at the second phase is relatively high and can be quickly reduced further after updating the dataset.} 
\label{fig::nav_training_logs}
\vspace{-5mm}
\end{figure}

We train SN-CBF models in two steps: initial training, and boundary refinement. 
The first step uses trajectory samples to roughly mark the safe and unsafe regions, and the second step focuses on sampling around the safety boundary from the first step, to refine it and improve its invariance properties. 
Both steps are important, as shown in Figure~\ref{fig::nav_training_logs}. 
The first step proposes safety boundaries from demonstrations to reduce the sampling space, and the second one corrects the values of misclassified states around the safety boundary. Both steps are performed in environments with a small number of obstacles, but will be deployed in much denser environments. 


\noindent{1)} {\em Initial Training.} 
We first collect a set of random trajectories of the robot interacting with the dynamic obstacles. This step can use a nominal simple controller with a high collision rate, such as a simple potential-field controller or an RL-trained reactive control policy. 
From these trajectories, we collect the initial labeling of safe states and unsafe states between the ego-agent and an obstacle based on whether collision occurs. 
For each state, we keep track of $h\in H$ that encodes the sequence of relative states between the robot and one obstacle. Thus we obtain an initial safe set $D_s\subseteq X\times H$ of collision-free samples, and an initial unsafe set $D_u\subseteq X\times H$ of samples in collision. These samples are sparse, and the initial training only relies on this weak supervision to approximately separate safe and unsafe regions.

Using the safe set $D_s$ and unsafe set $D_u$ of pairs $(x,h)$ collected through the demonstrations, we train the SN-CBF model by minimizing the following loss function, which encodes the standard CBF conditions (Definition~\ref{def:barrier_func}), with an error margin parameterized by $\gamma\in\mathbb{R}^+$:
\begin{eqnarray}
\label{eqn::barrier_objective}
    L_{B,D} &=&\frac{1}{N_{s}} \sum_{\small (x, h) \in D_s} \phi_{\gamma}(-B(x, h))\\
    &+&\frac{1}{N_{u}} \sum_{(x, h) \in D_u} \phi_{\gamma}(B(x, h))\\
         &+& \frac{1}{N} \sum_{(x, h) \in D} \phi_{\gamma}(-\dot{B}(x, h)-\alpha(B(x, h)))
\end{eqnarray}
where $\phi_{\gamma}(x) = \max(\gamma+x, 0)$.
The first term enforces that the the value of $B(x,h)$ for any safe $(x,h)\in D_s$
should be greater than $\gamma$, because a positive loss is only incurred when $\gamma-B(x,h)>0$. The second term 
enforces $B(x,h)$ to take sufficiently negative values on unsafe pairs. The third term enforces the Lie derivative condition $\dot B(x,h)\geq -\alpha(B(x,h))+\gamma$, where $\alpha$ is chosen to be a positive constant as an extended class $\mathcal{K}_{\infty}$ function. 
Because of the unknown interaction dynamics, the Lie derivative $\dot B$ can not be analytically computed, but can be approximated by the finite difference between two consecutive pairs, i.e., $\dot B(x,h)\dot=(B(x',h') - B(x,h))/\Delta t$.
The margin $\gamma$ is used to enforce the invariance conditions of CBFs.

\noindent{2)} {\em Boundary Refinement.} After the initial training, the SN-CBF models may violate the control barrier conditions in Definition~\ref{def:barrier_func} at many states near the safety boundary (i.e., the zero-levelset of the model). We then refine the model by focusing the training at this boundary between the safe and unsafe regions in the following steps. 





We first collect from the demonstrations from the previous step, as the initial set $D^{\theta}$ of $(x,h)$ pairs that are close to the safety boundary and currently classified as ``safe'' by the SN-CBF model obtained from the initial training. 
We then examine all elements in $D^{\theta}$. First, if some $(x,h)$ pair is already in collision and thus wrongly classified by the initial model, we remove it from $D_s$ and add it to $D_u$. Second, we examine the invariance condition on each pair by sampling control actions and take one that maximizes the predicted next state. This operation is an approximation of the $\max_{u\in U}$ operator in the CBF conditions in Definition~\ref{def:barrier_func}. 
We then inspect if the next state $x'$ under the best sampled action can be in collision.
If so, we add both $(x,h)$ and $(x',h')$ into $D_u$, where $h'$ is the corresponding new state sequence of the obstacle induced by this control action. 
After updating the $D_s$ and $D_u$, we retrain the SN-CBF models, still using (\ref{eqn::barrier_objective}). We iteratively perform this refinement until convergence.

\section{Online Inference with SN-CBF}
\label{section::planning}

After training the SN-CBF models for individual obstacles, we can apply them to an arbitrary number of obstacles individually, and the aggregate all values as follows:
\begin{equation}
    \mathcal B(x)=\prod_{i=1}^q \max\bigg(\frac{1}{b}\min\Big(B(x,h_i),b\Big),0\bigg)
\end{equation}
where $b\in \mathbb{R}^+$ is a threshold parameter. This aggregated $\mathcal{B}(x)$ value defines the total value landscape for the state $x$ of the ego-robot. This aggregation rule ensures that if $B(x,h_i)\leq 0$ for any obstacle $i$, then $\mathcal{B}(x)=0$ and the state $x$ is considered unsafe. On the other hand, $B(x,h_i)$ is clipped at $b$ for all $i$, so obstacles that are far from the ego-robot will not affect $\mathcal{B}(x)$. Overall, $x$ is unsafe with respect to any obstacle if and only if $\mathcal{B}(x)=0$, and $\mathcal{B}(x)$ is always within $[0,1]$. 



Using the aggregated $\mathcal{B}$ values, we compute control actions at each state $x$ of the ego-robot. We sample from the control action space $U$ for a fixed number of candidate control actions $u_1,...,u_l$. We then use the (learned) dynamics model of the ego-robot to predict its next state $x'_i=\pi(x,u_i)$ for each sampled action $u_i$, and evaluate the predicted next states $x'_i$ by $\mathcal{B}(x'_i)$. Any $u_i$ that corresponds to a nonzero $\mathcal{B}(x'_i)$ is considered a feasible action that can avoid collision. We then choose $u_i$ that corresponds to the next state that minimizes the distance between $x'_i$ and the goal. When no feasible action is available, we declare failure, and stop the robot. 

\begin{figure*}[h!]
\centering
\includegraphics[width=0.32\textwidth, height = 4cm]{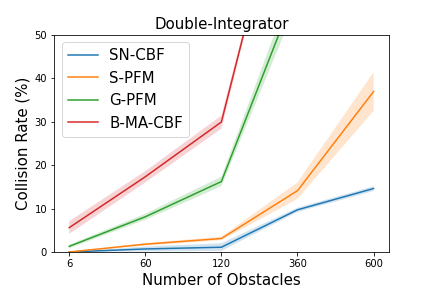}
\includegraphics[width=0.32\textwidth, height = 4cm]{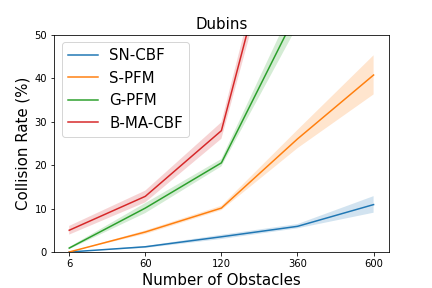}
\includegraphics[width=0.32\textwidth, height = 4cm]{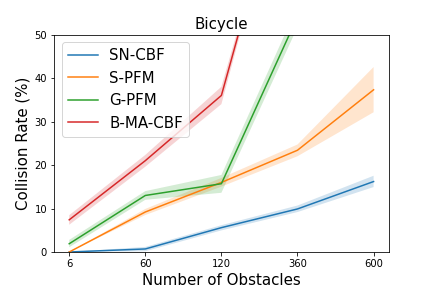}
\caption{\small Overall performance in the navigation environment using different underlying dynamics for the ego-robot. We measure the collision rate of our method (SN-CBF) and the other baselines, while scaling up the density of the dynamic obstacles. The underlying dynamics is stochastic and all agents are randomly initialized.  The shaded area shows variance over 5 random seeds.
}
\label{fig::nav_collision_rates}
\end{figure*}

\begin{figure*}[th!]
\centering
\includegraphics[width=0.25\textwidth]{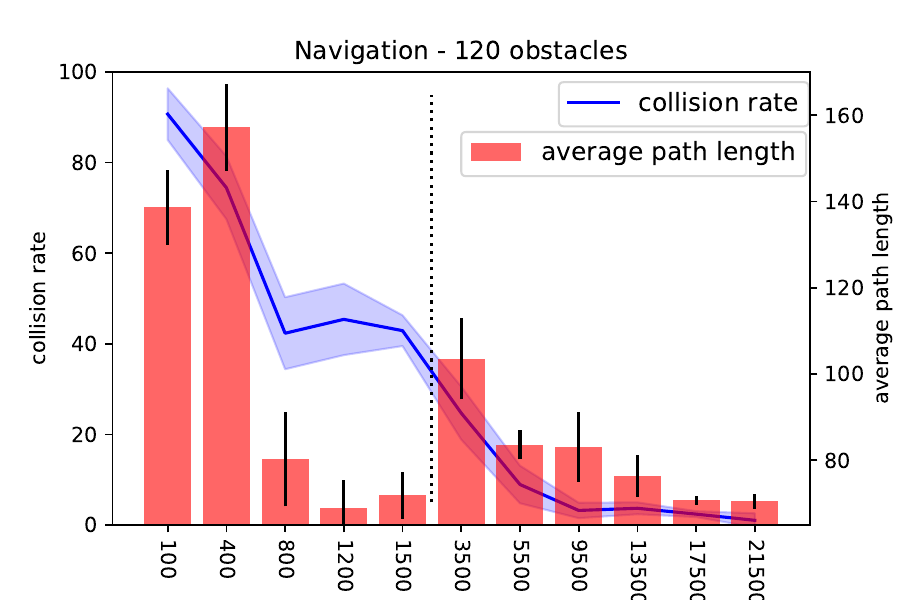}
\includegraphics[width=0.48\textwidth]{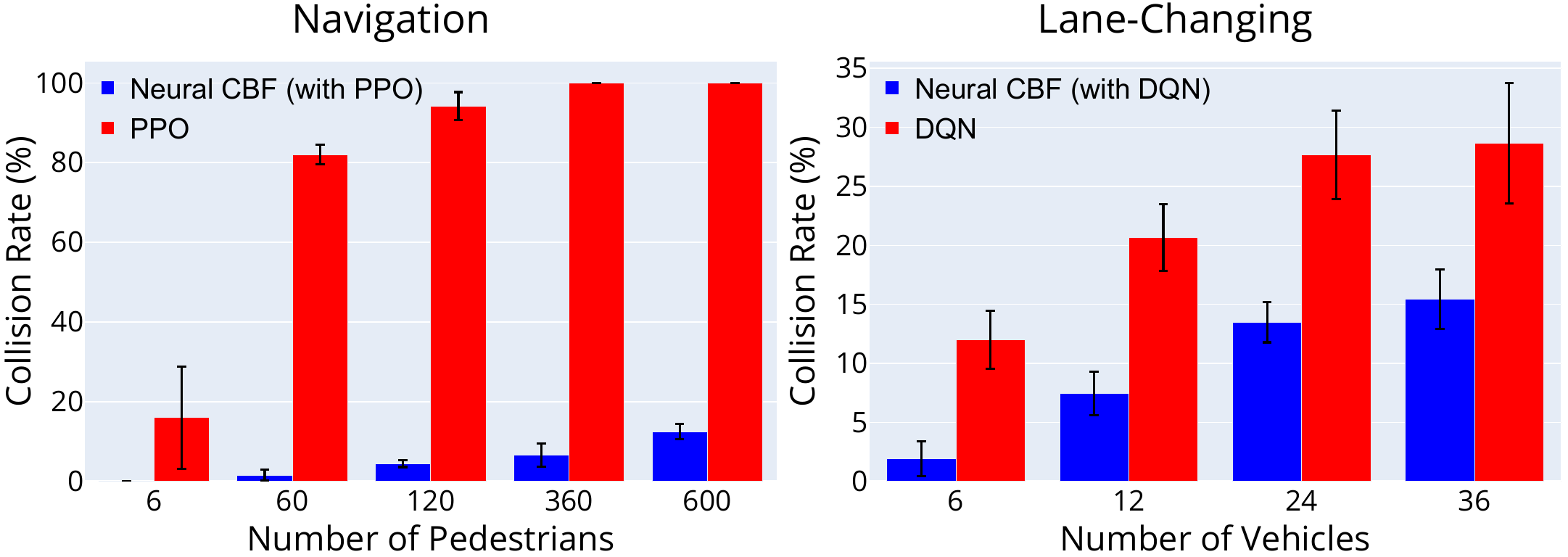}
\includegraphics[width=0.25\textwidth]{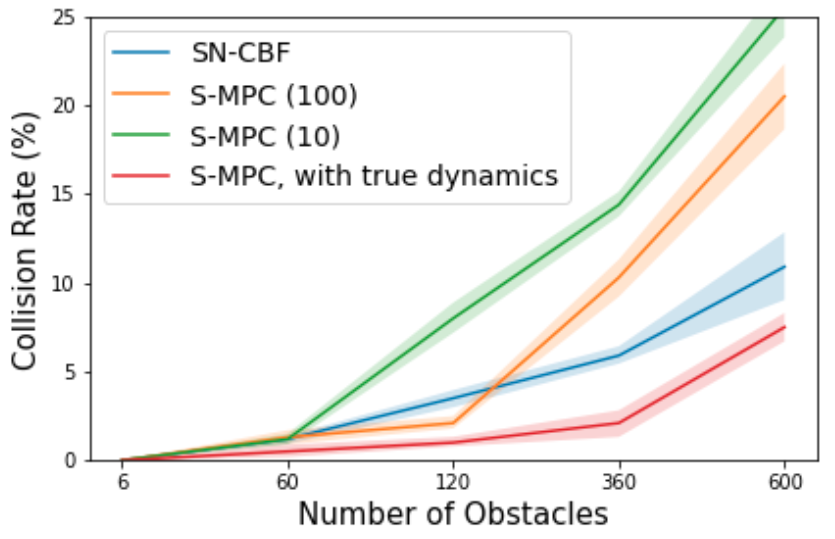}
\caption{{(Plot 1)} Test performance improves as training progresses. The vertical dashed line marks the start of boundary refinement procedures. {(Plot 2-3)} Comparison with reinforcement learning methods in the navigation environment and the highway lane-changing environment. {(Plot 4)} Comparisons with various settings of the sampling-based model predictive control (S-MPC) in the navigation environment with with Dubins car model for the ego-vehicle.} 
\label{fig:rl}\vspace{-.3cm}
\end{figure*}

\section{Experiments}\label{section::experiment}
We evaluate the proposed SN-CBF methods both in simulation and in hardware experiments. In simulation, we consider a robot navigation around pedestrians environment that can be easily scaled, as well as a highway lane-changing environment. In hardware experiments, we use SN-CBF to control directly an ego-robot car navigating around densely distributed pedestrians. The hardware experiment setting is shown in Figure~\ref{fig::unity} as well as the supplementary video. 

In the simulation environments, the pedestrians are modeled using the optimal reciprocal collision avoidance (ORCA) model~\cite{ORCA} and the vehicles on highway are modeled with the intelligent driver model (IDM)~\cite{treiber2000congested}. 
These underlying models are unknown to the learning agents. We test the methods with different densities of obstacles and different dynamics of the ego-robot, including single and double integrator, the Dubins car model, and the bicycle model.

\noindent{\bf Baselines Methods.} 
We adapt various existing methods into data-driven and sampling-based forms, and maintain their core approaches. We consider the following baselines:

- {\em Sampling-based potential field methods (S-PFM)}: a standard potential field method~\cite{potential86} with repulsive fields around each obstacle and attractive field around the goal based on Euclidean distance. In each step we sample actions and evaluate the predicted next states on these actions. 

- {\em Gradient-based potential field methods (G-PFM)}: a similar potential field method that uses gradient-based control based on the gradients of the potential fields. Note that it requires full knowledge of the dynamics of the ego-robot. 

- {\em Sampling-based MPC (S-MPC)}: a method that learns a neural dynamics model, unrolls the model online to construct a tree of future states, then selects the first action that leads to the best predicted outcome~\cite{stenman1999model,pan2011model}. 

- {\em Black-box multi-agent-CBF (B-MA-CBF)}: a method for safe multi-agent control that learns decentralized CBFs using known system dynamics~\cite{macbf}. The approach uses the max pooling layer design in neural network architecture instead of sequential modeling. 

- {\em Proximal policy optimization (PPO) and deep Q-Learning (DQN)}: two deep reinforcement learning methods, we use PPO~\cite{schulman2017proximal} for the continuous action space in the navigation environment, and DQN~\cite{mnih2013playing} for the discrete actions in the highway lane-changing environment. 






\noindent{\bf Simulation Experiment Setup.} In all evaluation experiments, we randomly initialize the agent, obstacles, and the goal configurations. We label a full trajectory as collision-free only when the agent successfully reaches the goal, with no collision or failure of finding control online at {\em any} step. Otherwise we consider the full trajectory as a failure. We define the collision rate to be the ratio of failed {\em trajectories} over the total. All experiments use 5 different random seeds. 



\noindent{\bf Hardware Experiment Setup.}
We train SN-CBF models for controlling a car robot to avoid pedestrians in an indoor environment. We first collect data from a small number of pedestrians, and adapt the ORCA model to provide a simulation model of the pedestrians. 
We perform the training procedures of SN-CBF models in simulation, and then deploy the SN-CBF models in the hardware car robot to infer control actions in real-time. 
We deploy the car in test environments with 
3 times the pedestrian density compared to the data collection phase, as shown in Figure~\ref{fig::unity}. 
We demonstrate the success of the methods in the supplementary video. 


\noindent{\bf Overall Performance Compared to Baselines.}
First, Figure \ref{fig::nav_collision_rates} compares the performance of SN-CBF 
for reducing collision in the simulation environment of navigation. 
The training is performed in simulation environments with only 6 obstacles, and the results show how the performance of the learned models scale as the density of obstacles increases up to 100 times of the training environment. The results confirm that sampling-based control outperforms gradient-based control (which assumes additional knowledge of the dynamics), especially when the environment becomes dense. 

When the ego-robot has simple dynamics that are easy to control, such as in the case of the single-integrator, the sampling-based potential field methods can perform quite well, but the gap with neural CBF becomes much larger in non-holonomic cases such as the Dubins car model. In all environments, SN-CBF
reduces the collision rate by more than 50\% from the best performing potential field methods. Across all environments, SN-CBF methods are able to maintain collision rate under 10\% up to 60x more obstacles, and only reach 15\% in the bicycle model case with 600 obstacles.

\begin{figure}[t]
\centering
\includegraphics[width=0.15\textwidth]{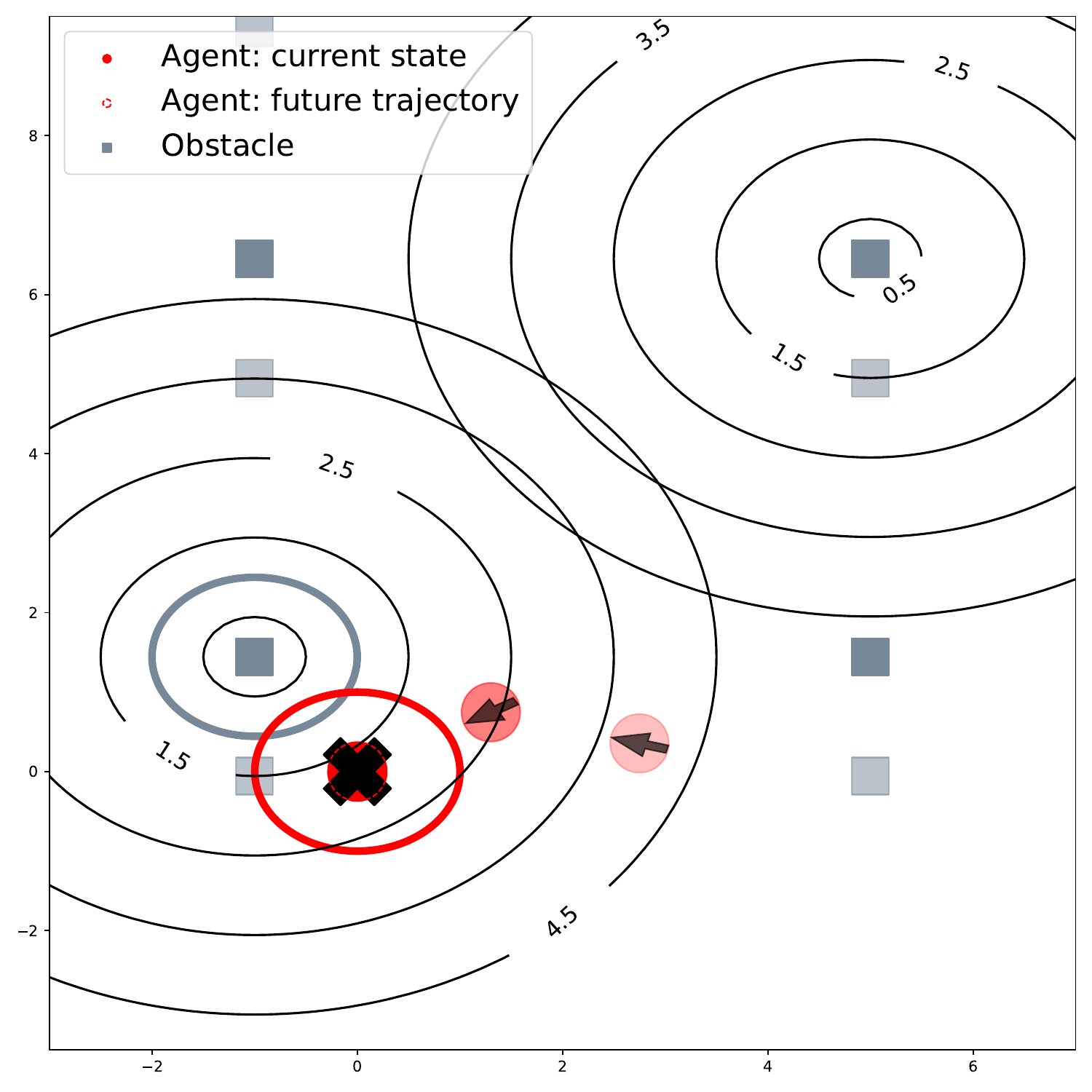}
\includegraphics[width=0.15\textwidth]{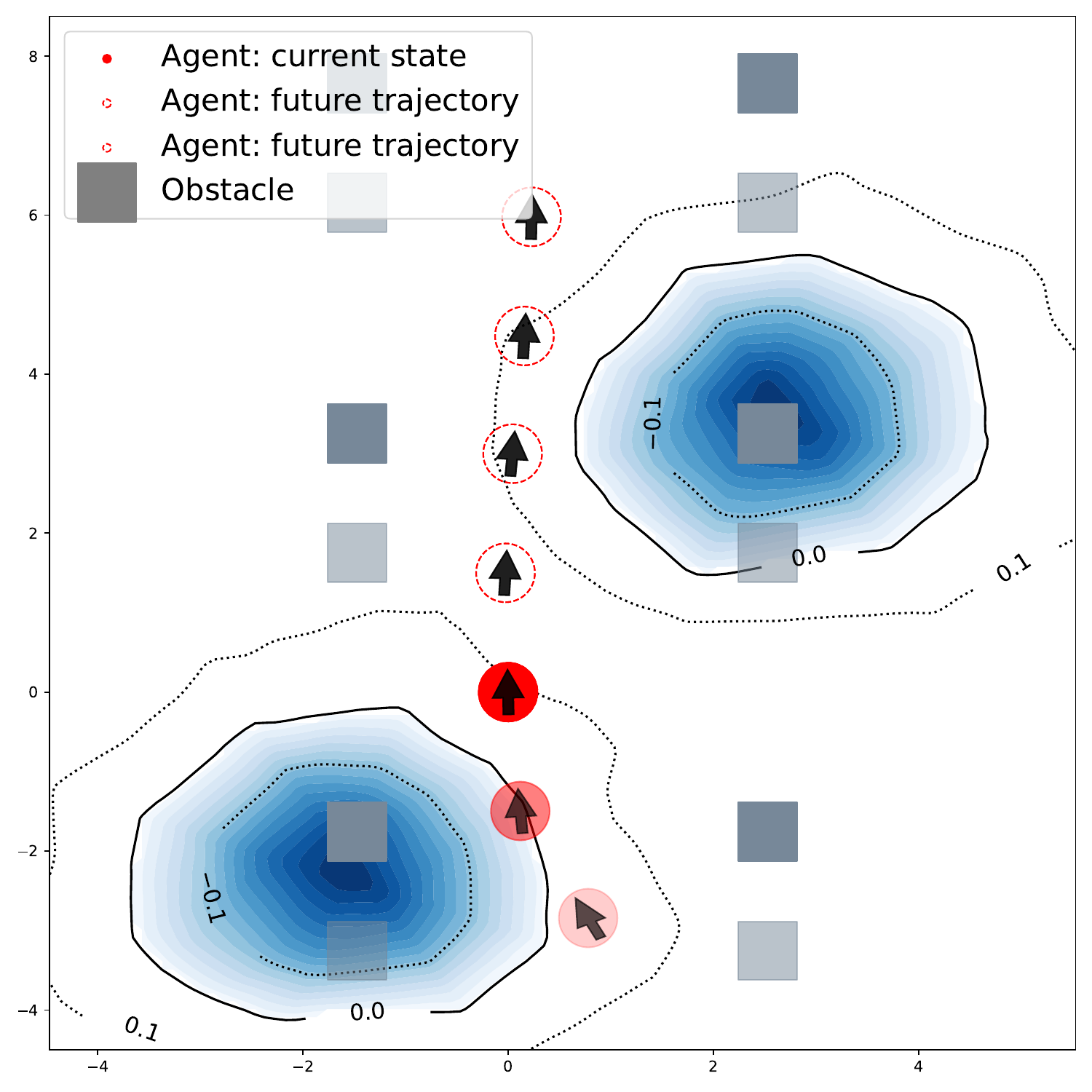}
\includegraphics[width=0.15\textwidth]{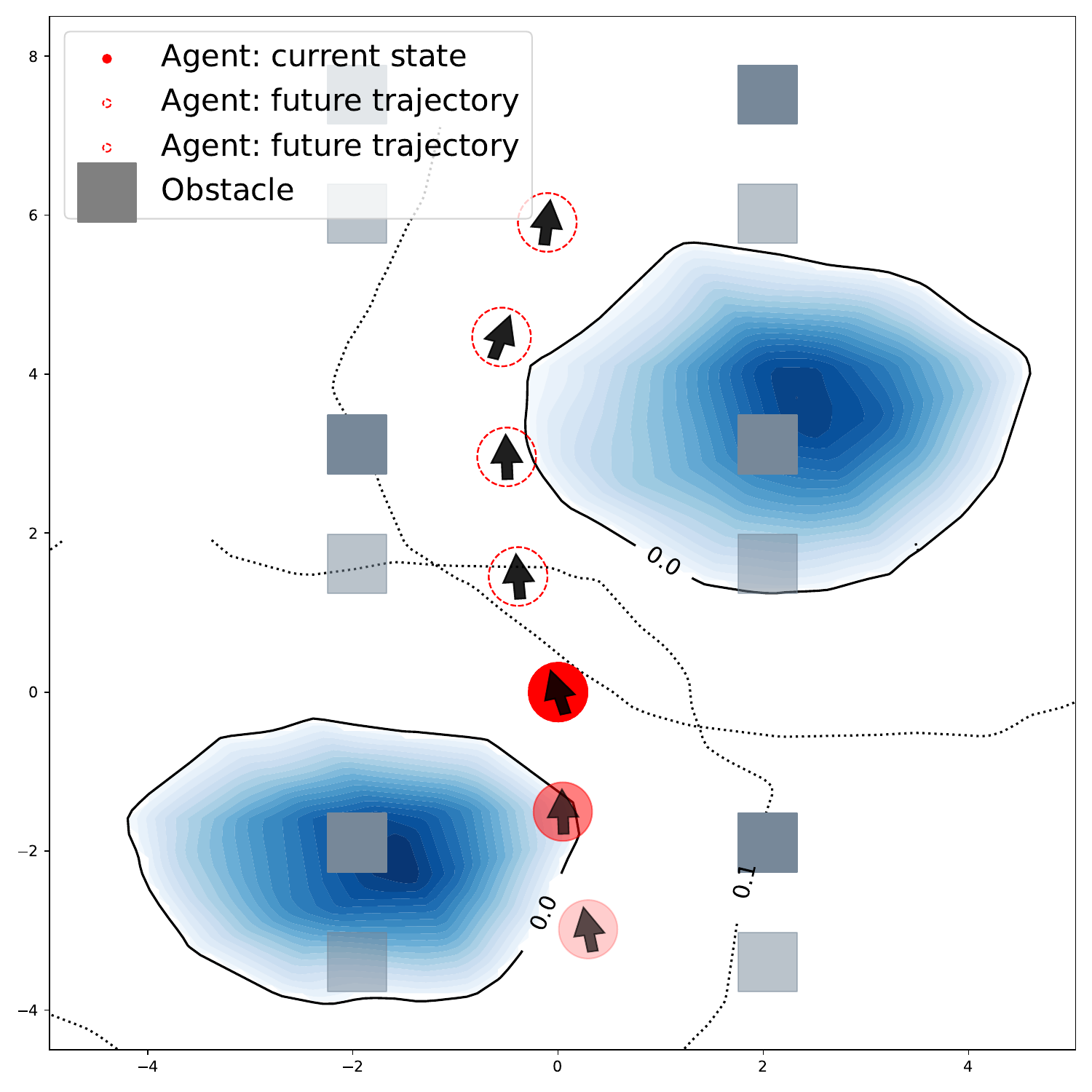}

\caption{\small {(Plot 1)} The ego-robot is controlled using gradient-based potential field methods, and it collides with an obstacle. {(Plot 2-3)} In both plots, the obstacle settings are the same and the ego-robot is able to pass through the corridor efficiently. The ego-robot uses the Dubins car dynamics in Plot 2, and the Bicycle model in Plot 3. We see that the CBF landscapes are quite different in the two cases, but the ego-robot is controlled well in both cases.}  
\label{fig::narrow_corridor}
\vspace{-.3cm}
\end{figure}


The comparison with B-MA-CBF confirms the importance of the sequential modeling choice. Note that this method is an adaptation of the original MA-CBF~\cite{macbf} to the model-free setting, so its generalizability becomes worse than the original training with known dynamics. The main factor for the performance difference is that MA-CBF uses an aggregation model on the spatial patterns of the adjacent obstacles, which enables it to handle a varying number of obstacles but the distribution drift in the spatial interaction patterns restricts generalization of the learned models. 


\noindent{\bf Comparison with End-to-End Reinforcement Learning.} Figure~\ref{fig:rl} shows the comparison with standard RL methods. In the navigation environment, the policy trained with PPO can perform reasonably in the training environment, but almost always fails in denser environments (collision rate reaching 100\% quickly). We use a version of SN-CBF methods that uses the control policy learned in PPO to provide the nominal control action for fair comparison, and we see that the collision reduction is still significant. In the lane-changing environment, we discretize the action space so that the comparison can be made with Deep Q-learning. This environment can not be made arbitrarily dense, and we still observe significant collision reduction. 

\noindent{\bf Alleviating Narrow Corridors in APF.}
The narrow corridor problem is a well-known issue in potential field methods~\cite{potential86}. When the ego-robot enters an area with where the adjacent obstacles create repulsive fields that point at conflicting directions, the robot can be misguided into collision or oscillation loops. In Figure~\ref{fig::narrow_corridor} we illustrate this problem where the collision case follows from gradient-based potential field control. 
In contrast, SN-CBF methods generate more accurate and dynamics-aware force fields to improve online control. 
In 
Figure~\ref{fig::narrow_corridor}
we show the level sets of the learned models for both the Dubins car model and the bicycle model. The different dynamics induced very different landscapes. In particular, the SN-CBF model in the bicycle case induces a much wider gap between the level sets, which reflects the need to initiate collision avoidance much farther away from the obstacles. In both cases, the dynamics-aware SN-CBF enables online control that maintains efficient movement to pass the corridor.

\noindent{\bf Comparison with Model-Predictive Approaches.}\label{compare_mpc} The standard setting of MPC requires the use of analytic dynamics of both the ego-robot and the obstacles, and thus can not be directly applied to the model-free setting. Instead, we can compare with a sampling-based adaptation of MPC by sampling control actions and forward predicting the future states, and then selecting control actions based on the potential field values of predicted states. This comparison allows us to understand the effectiveness of the SN-CBF models in capturing the dynamics without multi-step unrolling. 
In Figure~\ref{fig:rl} (Plot 4), 
we observe the benefits of CBF models in capturing the dynamic nature of the interactions through the barrier landscapes and avoid expensive online computation. 
It also allows us to avoid the accumulation of model-prediction errors that are inherent in learned models of dynamics. 



\section{Conclusion}

We proposed novel learning-based control methods for scalable dynamic obstacle avoidance through compositional learning of SN-CBF models. 
We exploit the important observation that the spatial interaction patterns of multiple obstacles can be decomposed and predicted through sequential modeling of individual obstacles. 
We design SN-CBF models that incorporate sequential modeling of individual obstacles, so that they can be composed in environments with an arbitrary number of obstacles. 
The online inference composes SN-CBF models of all the dynamic obstacles simultaneously to reduce the ``freezing the robot'' problem. We evaluated the methods by training in environments with a small number of obstacles, and tested the effectiveness of online composition and control in environments where the obstacle density is up to 100x higher. 
We have demonstrated the benefits in comparison with potential field methods, reinforcement learning, and sampling-based model-predictive approaches.
We believe SN-CBF methods can provide a powerful framework for tackling many challenging problems in robot control in the model-free settings. One direction for future work is the analysis of the probabilistic safety properties of the methods under certain assumptions on the environments.  

\noindent
\textbf{Acknowledgement.} 
The work is supported by NSF Career CCF 2047034, NSF AI Institute CCF 2112665, Amazon Research Award, and ONR YIP N00014-22-1-2292.




\bibliographystyle{unsrt}
\bibliography{references, sylvia_references}

\begin{thebibliography}{10}

\bibitem{potential86}
Oussama Khatib.
\newblock Real-time obstacle avoidance for manipulators and mobile robots.
\newblock In {\em Autonomous robot vehicles}, pages 396--404. Springer, 1986.

\bibitem{fox1997dynamic}
Dieter Fox, Wolfram Burgard, and Sebastian Thrun.
\newblock The dynamic window approach to collision avoidance.
\newblock {\em IEEE Robotics \& Automation Magazine}, 4(1):23--33, 1997.

\bibitem{kouvaritakis2016model}
Basil Kouvaritakis and Mark Cannon.
\newblock Model predictive control.
\newblock {\em Switzerland: Springer International Publishing}, page~38, 2016.

\bibitem{ji2016path}
Jie Ji, Amir Khajepour, Wael~William Melek, and Yanjun Huang.
\newblock Path planning and tracking for vehicle collision avoidance based on
  model predictive control with multiconstraints.
\newblock {\em IEEE Transactions on Vehicular Technology}, 66(2):952--964,
  2016.

\bibitem{Ames2019CBFOverview}
Aaron~D Ames, Samuel Coogan, Magnus Egerstedt, Gennaro Notomista, Koushil
  Sreenath, and Paulo Tabuada.
\newblock {Control barrier functions: Theory and applications}.
\newblock In {\em 2019 18th European Control Conference (ECC)}, pages
  3420--3431. IEEE, 2019.

\bibitem{Singletary2020APFCBF}
Andrew Singletary, Karl Klingebiel, Joseph Bourne, Andrew Browning, Phil
  Tokumaru, and Aaron Ames.
\newblock {Comparative analysis of control barrier functions and artificial
  potential fields for obstacle avoidance}.
\newblock {\em arXiv preprint arXiv:2010.09819}, 2020.

\bibitem{xiao2019control}
Wei Xiao and Calin Belta.
\newblock Control barrier functions for systems with high relative degree.
\newblock In {\em 2019 IEEE 58th conference on decision and control (CDC)},
  pages 474--479. IEEE, 2019.

\bibitem{huang2020switched}
Yiwen Huang and Yan Chen.
\newblock Switched control barrier function with applications to vehicle safety
  control.
\newblock In {\em Dynamic Systems and Control Conference}, volume 84270, page
  V001T15A002. American Society of Mechanical Engineers, 2020.

\bibitem{breeden2021robust}
Joseph Breeden and Dimitra Panagou.
\newblock Robust control barrier functions under high relative degree and input
  constraints for satellite trajectories.
\newblock {\em arXiv preprint arXiv:2107.04094}, 2021.

\bibitem{Fridovich2020HumanJournal}
David Fridovich-Keil, Andrea Bajcsy, Jaime~F. Fisac, Sylvia~L. Herbert, Steven
  Wang, Anca~D. Dragan, and Claire~J. Tomlin.
\newblock {Confidence-aware motion prediction for real-time collision
  avoidance}.
\newblock {\em International Journal of Robotics Research (IJRR)}, 2020.

\bibitem{freezingrobot}
Peter Trautman and Andreas Krause.
\newblock Unfreezing the robot: Navigation in dense, interacting crowds.
\newblock In {\em 2010 IEEE/RSJ International Conference on Intelligent Robots
  and Systems}, pages 797--803, 2010.

\bibitem{rudenko2020human}
Andrey Rudenko, Luigi Palmieri, Michael Herman, Kris~M Kitani, Dariu~M Gavrila,
  and Kai~O Arras.
\newblock Human motion trajectory prediction: A survey.
\newblock {\em The International Journal of Robotics Research}, 39(8):895--935,
  2020.

\bibitem{Kruse2013HumanAwareNav}
Thibault Kruse, Amit~Kumar Pandey, Rachid Alami, and Alexandra Kirsch.
\newblock {Human-aware robot navigation: A survey}.
\newblock {\em Robotics and Autonomous Systems}, 61(12):1726--1743, 2013.

\bibitem{willard2020integrating}
Jared Willard, Xiaowei Jia, Shaoming Xu, Michael Steinbach, and Vipin Kumar.
\newblock Integrating physics-based modeling with machine learning: A survey.
\newblock {\em arXiv preprint arXiv:2003.04919}, 1(1):1--34, 2020.

\bibitem{kocijan2016modelling}
Ju{\v{s}} Kocijan.
\newblock {\em Modelling and control of dynamic systems using Gaussian process
  models}.
\newblock Springer, 2016.

\bibitem{nair2018overcoming}
Ashvin Nair, Bob McGrew, Marcin Andrychowicz, Wojciech Zaremba, and Pieter
  Abbeel.
\newblock Overcoming exploration in reinforcement learning with demonstrations.
\newblock In {\em 2018 IEEE International Conference on Robotics and Automation
  (ICRA)}, pages 6292--6299. IEEE, 2018.

\bibitem{yang2020reinforcement}
Lin Yang and Mengdi Wang.
\newblock Reinforcement learning in feature space: Matrix bandit, kernels, and
  regret bound.
\newblock In {\em International Conference on Machine Learning}, pages
  10746--10756. PMLR, 2020.

\bibitem{Barraquand1992APF}
Jerome Barraquand, Bruno Langlois, and J-C Latombe.
\newblock {Numerical potential field techniques for robot path planning}.
\newblock {\em IEEE transactions on systems, man, and cybernetics},
  22(2):224--241, 1992.

\bibitem{Ge2002APF}
Shuzhi~Sam Ge and Yun~J Cui.
\newblock {Dynamic motion planning for mobile robots using potential field
  method}.
\newblock {\em Autonomous robots}, 13(3):207--222, 2002.

\bibitem{Henderson1971Fluid}
L~F Henderson.
\newblock {The statistics of crowd fluids}.
\newblock {\em nature}, 229(5284):381--383, 1971.

\bibitem{Rudenko2020HumanPredictionSurvey}
Andrey Rudenko, Luigi Palmieri, Michael Herman, Kris~M Kitani, Dariu~M Gavrila,
  and Kai~O Arras.
\newblock {Human motion trajectory prediction: A survey}.
\newblock {\em The International Journal of Robotics Research}, 39(8):895--935,
  2020.

\bibitem{bemporad1999robust}
Alberto Bemporad and Manfred Morari.
\newblock Robust model predictive control: A survey.
\newblock In {\em Robustness in identification and control}, pages 207--226.
  Springer, 1999.

\bibitem{stenman1999model}
Anders Stenman.
\newblock Model-free predictive control.
\newblock In {\em Proceedings of the 38th IEEE Conference on Decision and
  Control (Cat. No. 99CH36304)}, volume~4, pages 3712--3717. IEEE, 1999.

\bibitem{pan2011model}
Yunpeng Pan and Jun Wang.
\newblock Model predictive control of unknown nonlinear dynamical systems based
  on recurrent neural networks.
\newblock {\em IEEE Transactions on Industrial Electronics}, 59(8):3089--3101,
  2011.

\bibitem{chen2017socially}
Yu~Fan Chen, Michael Everett, Miao Liu, and Jonathan~P How.
\newblock Socially aware motion planning with deep reinforcement learning.
\newblock In {\em 2017 IEEE/RSJ International Conference on Intelligent Robots
  and Systems (IROS)}, pages 1343--1350. IEEE, 2017.

\bibitem{long2018towards}
Pinxin Long, Tingxiang Fan, Xinyi Liao, Wenxi Liu, Hao Zhang, and Jia Pan.
\newblock Towards optimally decentralized multi-robot collision avoidance via
  deep reinforcement learning.
\newblock In {\em 2018 IEEE international conference on robotics and automation
  (ICRA)}, pages 6252--6259. IEEE, 2018.

\bibitem{everett2021collision}
Michael Everett, Yu~Fan Chen, and Jonathan~P How.
\newblock Collision avoidance in pedestrian-rich environments with deep
  reinforcement learning.
\newblock {\em IEEE Access}, 9:10357--10377, 2021.

\bibitem{ma2021model}
Haitong Ma, Jianyu Chen, Shengbo Eben, Ziyu Lin, Yang Guan, Yangang Ren, and
  Sifa Zheng.
\newblock Model-based constrained reinforcement learning using generalized
  control barrier function.
\newblock In {\em 2021 IEEE/RSJ International Conference on Intelligent Robots
  and Systems (IROS)}, pages 4552--4559. IEEE, 2021.

\bibitem{Gurriet2018CBFActiveSetInvariance}
Thomas Gurriet, Andrew Singletary, Jacob Reher, Laurent Ciarletta, Eric Feron,
  and Aaron Ames.
\newblock {Towards a framework for realizable safety critical control through
  active set invariance}.
\newblock In {\em 2018 ACM/IEEE 9th International Conference on Cyber-Physical
  Systems (ICCPS)}, pages 98--106. IEEE, 2018.

\bibitem{Nguyen2016CBFSteppingStones1}
Quan Nguyen, Ayonga Hereid, Jessy~W Grizzle, Aaron~D Ames, and Koushil
  Sreenath.
\newblock {3d dynamic walking on stepping stones with control barrier
  functions}.
\newblock In {\em 2016 IEEE 55th Conference on Decision and Control (CDC)},
  pages 827--834. IEEE, 2016.

\bibitem{Huang2020MultiAgentCBF}
Xinyuan Huang, Li~Li, and Jie Chen.
\newblock {Multi-agent system motion planning under temporal logic
  specifications and control barrier function}.
\newblock {\em Control Theory and Technology}, 18(3):269--278, 2020.

\bibitem{Srinivasan2018MulitAgentCBF}
Mohit Srinivasan, Samuel Coogan, and Magnus Egerstedt.
\newblock {Control of multi-agent systems with finite time control barrier
  certificates and temporal logic}.
\newblock In {\em 2018 IEEE Conference on Decision and Control (CDC)}, pages
  1991--1996. IEEE, 2018.

\bibitem{DBLP:conf/cdc/WangAE16}
Li~Wang, Aaron~D. Ames, and Magnus Egerstedt.
\newblock Multi-objective compositions for collision-free connectivity
  maintenance in teams of mobile robots.
\newblock In {\em 55th {IEEE} Conference on Decision and Control, {CDC} 2016,
  Las Vegas, NV, USA, December 12-14, 2016}, pages 2659--2664. {IEEE}, 2016.

\bibitem{DBLP:journals/trob/WangAE17}
Li~Wang, Aaron~D. Ames, and Magnus Egerstedt.
\newblock Safety barrier certificates for collisions-free multirobot systems.
\newblock {\em {IEEE} Trans. Robotics}, 33(3):661--674, 2017.

\bibitem{DBLP:journals/csysl/GlotfelterCE17}
Paul Glotfelter, Jorge Cort{\'{e}}s, and Magnus Egerstedt.
\newblock Nonsmooth barrier functions with applications to multi-robot systems.
\newblock {\em {IEEE} Control. Syst. Lett.}, 1(2):310--315, 2017.

\bibitem{DBLP:conf/ccta/GlotfelterCE18}
Paul Glotfelter, Jorge Cort{\'{e}}s, and Magnus Egerstedt.
\newblock Boolean composability of constraints and control synthesis for
  multi-robot systems via nonsmooth control barrier functions.
\newblock In {\em {IEEE} Conference on Control Technology and Applications,
  {CCTA} 2018, Copenhagen, Denmark, August 21-24, 2018}, pages 897--902.
  {IEEE}, 2018.

\bibitem{saveriano2019learning}
Matteo Saveriano and Dongheui Lee.
\newblock Learning barrier functions for constrained motion planning with
  dynamical systems.
\newblock In {\em IEEE International Conference on Intelligent Robots and
  Systems}, 2019.

\bibitem{Srinivasan2020LearnCBFSupervisedML}
Mohit Srinivasan, Amogh Dabholkar, Samuel Coogan, and Patricio~A Vela.
\newblock {Synthesis of control barrier functions using a supervised machine
  learning approach}.
\newblock In {\em 2020 IEEE/RSJ International Conference on Intelligent Robots
  and Systems (IROS)}, pages 7139--7145. IEEE, 2020.

\bibitem{Jin2020}
Wanxin Jin, Zhaoran Wang, Zhuoran Yang, and Shaoshuai Mou.
\newblock {Neural Certificates for Safe Control Policies}.
\newblock {\em arXiv}, jun 2020.

\bibitem{Taylor2020LearningCBF}
Andrew Taylor, Andrew Singletary, Yisong Yue, and Aaron Ames.
\newblock {Learning for safety-critical control with control barrier
  functions}.
\newblock In {\em Learning for Dynamics and Control}, pages 708--717. PMLR,
  2020.

\bibitem{Robey2020LearnCBFExpertDemonstrations}
Alexander Robey, Haimin Hu, Lars Lindemann, Hanwen Zhang, Dimos~V Dimarogonas,
  Stephen Tu, and Nikolai Matni.
\newblock {Learning control barrier functions from expert demonstrations}.
\newblock In {\em 2020 59th IEEE Conference on Decision and Control (CDC)},
  pages 3717--3724. IEEE, 2020.

\bibitem{macbf}
Zengyi Qin, Kaiqing Zhang, Yuxiao Chen, Jingkai Chen, and Chuchu Fan.
\newblock Learning safe multi-agent control with decentralized neural barrier
  certificates.
\newblock In {\em 9th International Conference on Learning Representations,
  {ICLR} 2021, Virtual Event, Austria, May 3-7, 2021}. OpenReview.net, 2021.

\bibitem{Dawson2021CBFNN}
Charles Dawson, Zengyi Qin, Sicun Gao, and Chuchu Fan.
\newblock {Safe Nonlinear Control Using Robust Neural Lyapunov-Barrier
  Functions}.
\newblock {\em arXiv preprint arXiv:2109.06697}, 2021.

\bibitem{ma2022learning}
Hengbo Ma, Bike Zhang, Masayoshi Tomizuka, and Koushil Sreenath.
\newblock Learning differentiable safety-critical control using control barrier
  functions for generalization to novel environments, 2022.

\bibitem{ORCA}
Jur Van Den~Berg, Stephen~J Guy, Ming Lin, and Dinesh Manocha.
\newblock Reciprocal n-body collision avoidance.
\newblock In {\em Robotics research}, pages 3--19. Springer, 2011.

\bibitem{hochreiter1997long}
Sepp Hochreiter and J{\"u}rgen Schmidhuber.
\newblock Long short-term memory.
\newblock {\em Neural computation}, 9(8):1735--1780, 1997.

\bibitem{treiber2000congested}
Martin Treiber, Ansgar Hennecke, and Dirk Helbing.
\newblock Congested traffic states in empirical observations and microscopic
  simulations.
\newblock {\em Physical review E}, 62(2):1805, 2000.

\bibitem{schulman2017proximal}
John Schulman, Filip Wolski, Prafulla Dhariwal, Alec Radford, and Oleg Klimov.
\newblock Proximal policy optimization algorithms.
\newblock {\em arXiv preprint arXiv:1707.06347}, 2017.

\bibitem{mnih2013playing}
Volodymyr Mnih, Koray Kavukcuoglu, David Silver, Alex Graves, Ioannis
  Antonoglou, Daan Wierstra, and Martin Riedmiller.
\newblock Playing atari with deep reinforcement learning.
\newblock {\em arXiv preprint arXiv:1312.5602}, 2013.

\end{thebibliography}


\end{document}

\begin{table*}[!bth]
\centering\small
\begin{tabular}{|c|c|c|c|c|}
\hline
 \diagbox{Method}{Vehicle Count}     & 6           & 12          & 24          & 36          \\ \hline
NCBF & 1.93 (0.24) & 7.43 (0.38) & 13.5 (0.33) & 15.4 (0.70) \\ \hline
DQN  & 12.0 (0.67) & 20.7 (0.89) & 27.7 (1.56) & 28.6 (2.89) \\ \hline
\end{tabular}
\caption{The over collision rates (\%) in the lane-changing environment.}
\label{table::lane-changing}
\end{table*}